\documentclass{elsarticle}
\usepackage{algorithm}
\usepackage{algpseudocode}
\usepackage{lineno}
\usepackage{diagbox}
\usepackage{booktabs}

\journal{A Nice Journal}

\usepackage{url}
\usepackage{hyperref}

\usepackage{graphicx} 
\usepackage{caption} 
\usepackage[skip=2pt,font=scriptsize]{subcaption}

\begin{document}

\begin{frontmatter}

\title{Quantitative Analysis of Particles Segregation}


\author[firstaddress]{Ting Peng}
\address[firstaddress]{MOE Key Laboratory of Special Area Highway Engineering, Chang'an University, Xi'an,710064 China}
\ead{t.peng@ieee.org}

\author[secondaddress]{Aiping Qu}
\address[secondaddress]{Shanxi Traffic Vocational and Technical College, Taiyuan, 030031 China}
\ead{562639350@qq.com}

\author[thirdaddress]{Xiaoling Wang}
\address[thirdaddress]{Xi'an Branch of the People's Bank of China, Xi'an, 710004,China}
\ead{xlwang\_xa@163.com}




\begin{abstract}

Segregation is a popular phenomenon. It has considerable effects on material performance. To the author's knowledge, there is still no automated objective Quantitative indicator for segregation. In order to full fill this task, segregation of particles is analyzed. Edges of the particles are extracted from the digital picture. Then, the whole picture of particles is splintered to small rectangles with the same shape. Statistical index of the edges in each rectangle is calculated. Accordingly, segregation between  the indexes corresponding to the rectangles is evaluated. The results show coincident with subjective evaluated results. Further more, it can be implemented as an automated system, which would facilitate the materials quality control mechanism during production process.
\end{abstract}


\end{frontmatter}


\newenvironment{vardesc}[1]{%
\settowidth{\parindent}{#1:\ }
\makebox[0pt][r]{#1:\ }}{}

\section{Introduction}

Segregation is a common phenomenon in nature. If it exists in society, it means higher probability of conflicts. If it exists in material, it means defects of materials. In materials, segregation almost always means decreasing of performance. In order to ensure reliable performance of materials, segregation should be evaluated objectively and quantitatively. Then, it can be controlled during material producing process. 

However, segregation is analyzed empirically in most research works currently. To the author's knowledge, there is no automated method available to evaluate segregation objectively and quantitatively. As far as we know, only  one objective evaluation index of segregation is available, in which segregation degree is evaluated according to weight difference between upper layer and lower layer\cite{Gao2015} of a specimen of concrete. It is a laborious process in practice.
Unfortunately,  there is still no automated method available to evaluate segregation objectively and quantitatively.  Because segregation has considerable influence in many areas, automated objective segregation evaluation method would be a perspective technique to ensure better performance in related areas.

Pattern recognition of digital picture provides a promising way to extract information automatically. It would be a nice method to full fill this task. In this region, different algorithms are proposed to conceal adverse effects of noise\cite{Setayesh2013}, to detect majority of variations in images\cite{Zareizadeh2013} and to detect edges with the influence of light\cite{Zhao2014}. A neutrosophic edge detection algorithm can remove the noise effect and detect the edges on both the noise-free images and the images with different levels of noises\cite{Guo2014}. New approaches improving feature extracting performance are proposing constantly\cite{Etemad2011,Gonzalez}.

Among them, fast edge detection using structured forests is an effective method to obtain edges from digital images.
It formulates the problem of predicting local edge masks in a structured learning framework applied to random decision forests\cite{6751339}, the structure present in local image patches is utilized to learn both an accurate and computationally efficient edge detector\cite{DollarARXIV14edges}. The number of contours that are wholly contained in a bounding box is indicative of the likelihood of the box containing an object\cite{ZitnickECCV14edgeBoxes}. The result is an approach that obtains real time performance that is orders
of magnitude faster than many competing state-of-the-art approaches\cite{6751339}.

In this work, an automated segregation evaluation method is proposed. At first, photo of the specimen is taken, edges of the particles are extracted from the picture with fast edge detection using structured forests\cite{6751339,DollarARXIV14edges,ZitnickECCV14edgeBoxes}. Then, the picture is splintered into parts with the same size. Segregation index is calculated according to the edges in each part. The experimental results show that the calculated index is correspond with empirically analyzing results. The whole process is easy to implement as an automated one.

The rest of the paper is structured as follows. We discuss segregation and its evaluation method in the next section. In this section, segregation index is proposed and the algorithm to compute it is also given. Experimental results and discussion are given in Section 3. At last, this work is concluded in Section 4.

\section{Proposed Strategy}

In this work, segregation is computed according to the digital picture of the particles.  At first, the picture of the particles is loaded. Then, edges of the particles are extracted with fast edge detection strategy\cite{DollarARXIV14edges}. In order to facilitate the processing speed, the picture is converted to black and white, extracted edges are shrinked. In order to evaluate segregation, the picture is splintered into $rows \times cols$ parts with the same size. Average edge length corresponding to each part is computed. At last, segregation between the edge length of the parts is calculated.
The detail of the strategy is shown in Algorithm \ref{alg:segregation}. 

\begin{algorithm}
\caption{Segregation Evaluation Process}\label{alg:segregation}
\begin{algorithmic}[2]
\Procedure{Segregation}{$picture,rows,cols$}
\Comment{The picture to be evaluated, it would be splintered to $rows \times cols$ parts with the same size}
   \State $pic \gets read\_picture(picture)$  \Comment{Load picture}
   \State $E \gets EdgeDetect(pic)$ \Comment{Extract edges from $pic$}
   \State $bw \gets pic2bw(E)$ \Comment{Convert to black and white}
   \State $respic \gets ShrinkEdge(bw)$ \Comment{Shrink the edges}
   \State $SplinteredPicture \gets SplitPicture(respic,rows,cols)$ \Comment{Split it into $rows \times cols$ parts}
   \For {$i \in range(rows)$}
       \For {$j in range(cols)$}
           \State $Res[i][j] \gets AverageEdgeLength(SplinteredPicture[i][j])$
       \EndFor
   \EndFor
   \State $SegregationIndex \gets MeasureSegregate(Res)$
   \State \textbf{return} $SegregationIndex$
\EndProcedure
\end{algorithmic}
\end{algorithm}

In the Algorithm \ref{alg:segregation}, segregation degree is calculated according to the following Equation \ref{eq:segregation}, which is  constructed according to Gini coefficient\cite{gini_coeff}.

\begin{equation} \label{eq:segregation} 
SegregateIndex = \frac{n+1-2\frac{\sum\limits_{i=1}^{n}{(n+1-i)y_i}}{\sum\limits_{i=1}^{n}{y_i}}}{n-1}
\end{equation}
\begin{vardesc}{Where}

$SegregateIndex$-- Segregate extend between the parts.

$n$-- Quantity of the parts.

$i$-- The index of the parts.

$y_i$-- Edge length of the $i$th part.

\end{vardesc}

In Equation \ref{eq:segregation}, $y_i$ stands for total length of the extracted edges in $i$th part. Segregation degree is calculated according to the difference of the edge length between the parts. If each part has the same edge length, the result will be zero. It means that no segregation exists between the parts.

If the edges of the whole picture are all concentrated in one part, the calculated result will be $1$. In this circumstance, segregation of the particles in the picture is at its highest point. 

In most cases, the extracted edges are distributed among the parts. The calculated value is in the interval of $(0,1)$. The bigger segregation index value means more serious segregation between the parts. 

In this way, segregation of the objects in the picture is converted to segregation between the edge length of the parts. Then, segregation degree is evaluated automatically.  

This process can be employed in evaluation of segregation between the particles. It is an objective evaluation method.  However, this method is also influenced by the relative size of the particles and the area size covered by the picture. The number of the parts also has some influence on the results. In order to obtain reliable results, the picture should cover enough particles. In this work, the number should be more than $20$. On the other hand, if the size of the picture is too big, the edges of the particles would uniformly distributed among the parts. In this circumstance, it would be hard to evaluate segregation. Accordingly, particles covered by the picture are less than three hundred in this work.

\section{Experimental Results and Discussion}

\subsection{Results}

\begin{figure}
\centering 
\begin{subfigure}[b]{0.23\textwidth} \includegraphics[width=\textwidth]{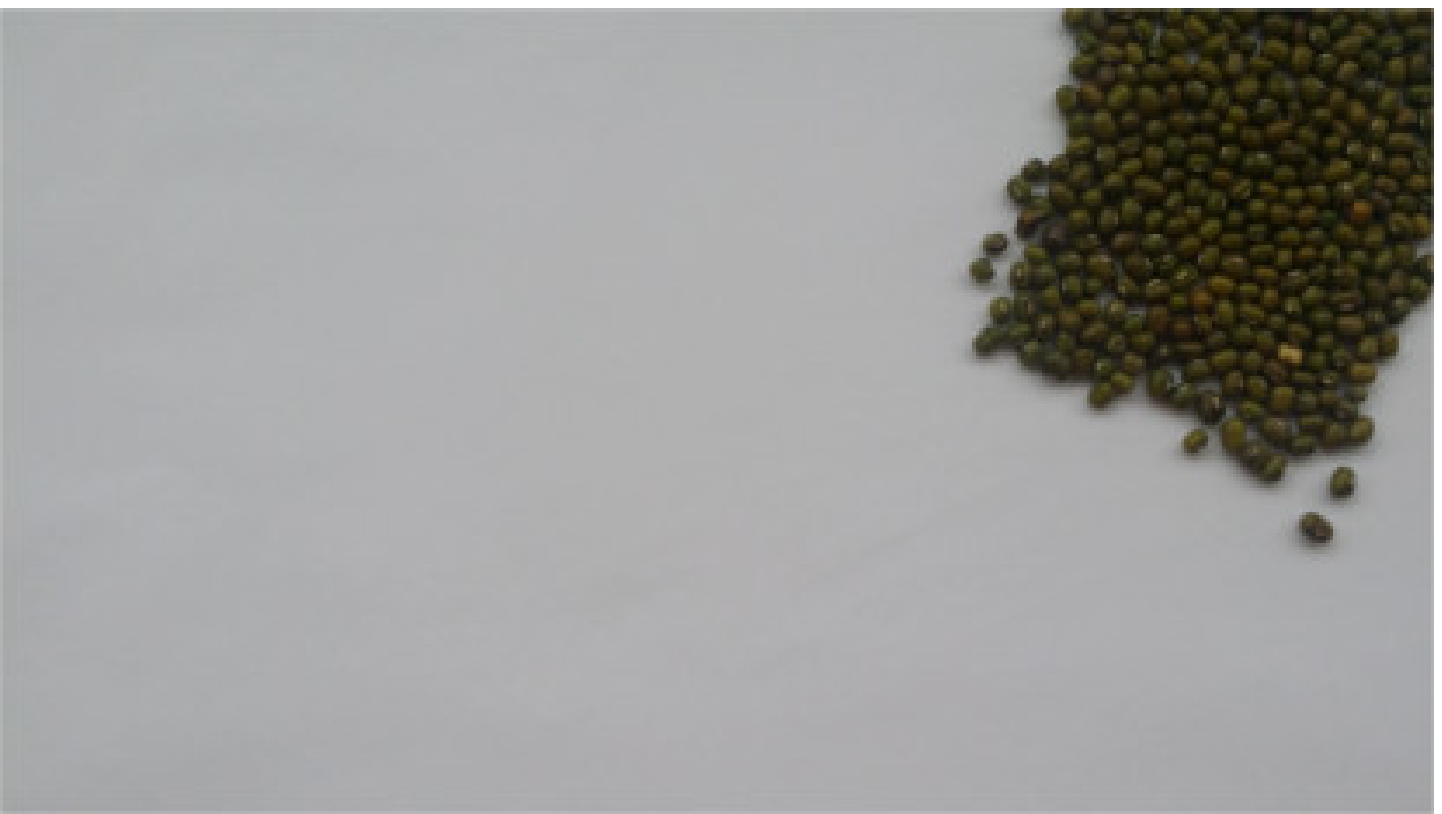} \caption{Sample $1$} \label{fig:mung1} \end{subfigure} ~ 
\begin{subfigure}[b]{0.23\textwidth} \includegraphics[width=\textwidth]{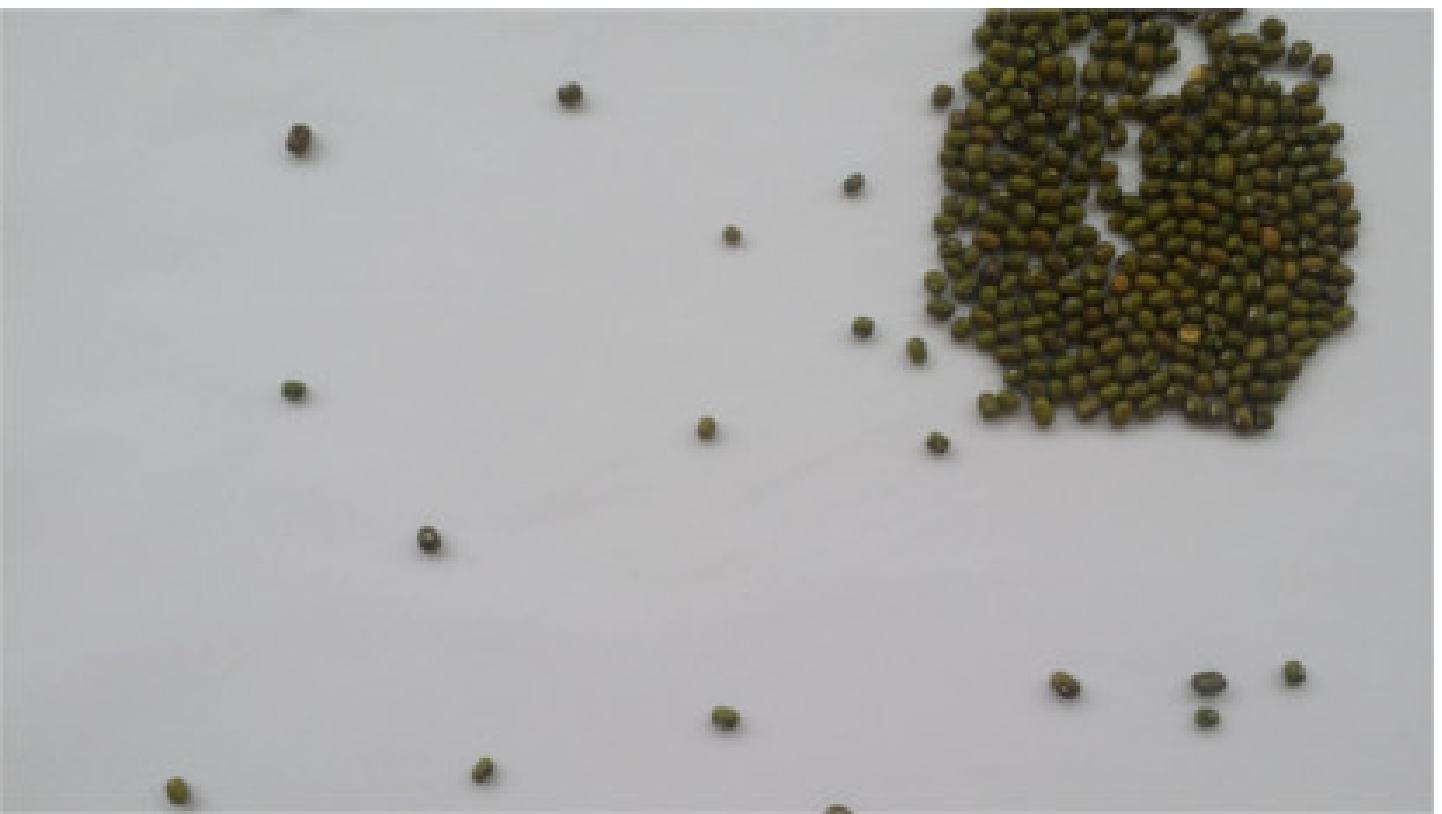} \caption{Sample $2$} \label{fig:mung2} \end{subfigure} ~ 
\begin{subfigure}[b]{0.23\textwidth} \includegraphics[width=\textwidth]{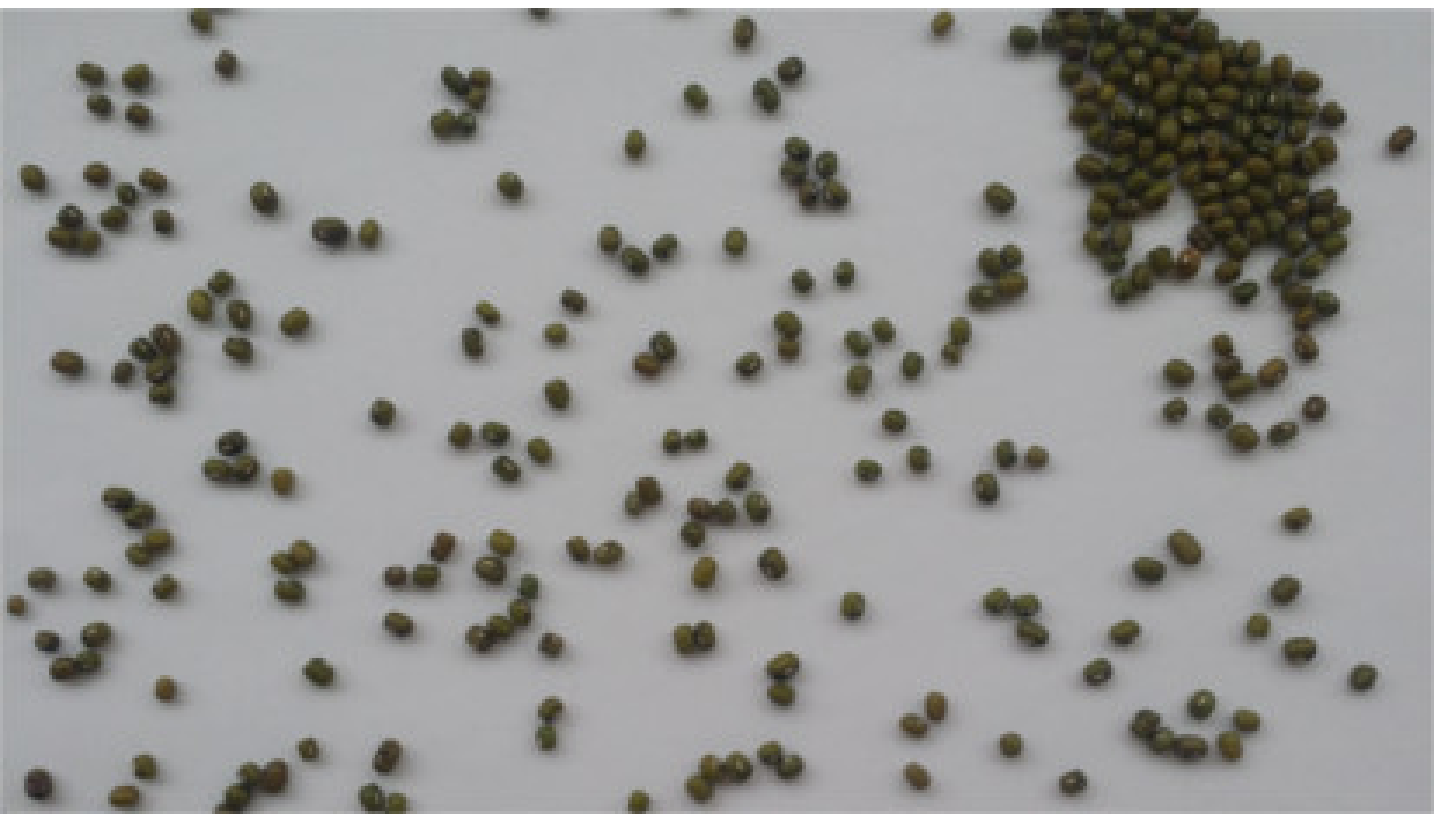} \caption{Sample $3$} \label{fig:mung3} \end{subfigure} ~
\begin{subfigure}[b]{0.23\textwidth} \includegraphics[width=\textwidth]{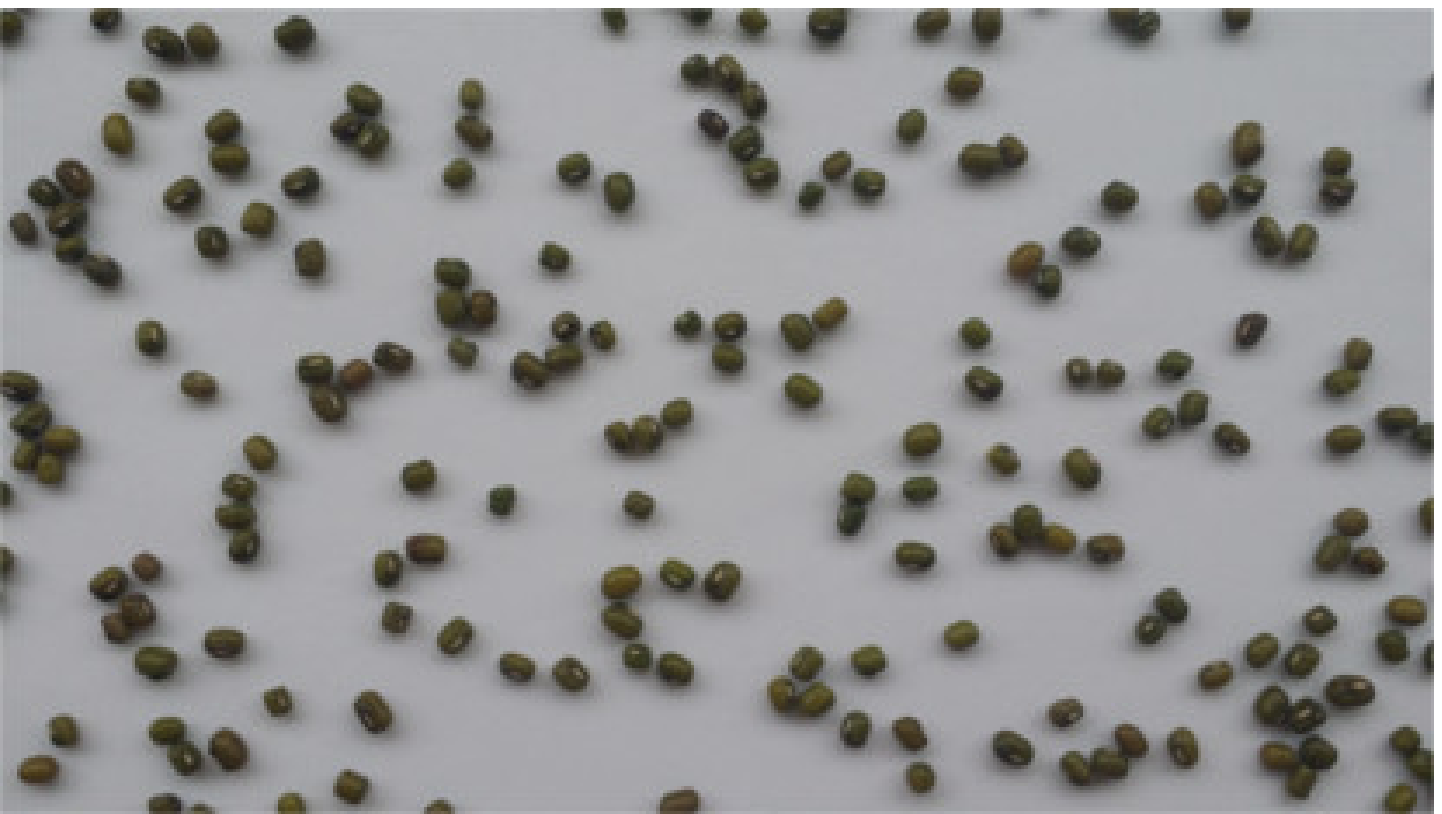} \caption{Sample $4$} \label{fig:mung4} \end{subfigure} 
\caption{Distribution of the Mungs}\label{fig:mungs} 
\end{figure}

\begin{table}
\scriptsize
\begin{center}
\caption{Segregation Result of Mungs(Sample $1$) }
\label{tab:mungA_res}
\begin{tabular}{c|cccccccc}
\toprule
\diagbox{Rows}{Cols} & 1 & 2 & 3 & 4 & 5 & 6 & 7 & 8  \\
\hline
1 &    0   &     1.0000  &   1.0000 &   0.9080  &  0.8500  &  0.8383  &  0.8172  &  0.8005 \\
2 &    0.8496 &   0.9499 &   0.9699 &   0.9176  &  0.8999  &  0.8992  &  0.8823  &  0.8787 \\ 
3 &    0.6164 &   0.8466 &   0.9041 &   0.8801  &  0.8595  &  0.8622  &  0.8603  &  0.8533 \\
4 &    0.6059 &   0.8311 &   0.8925 &   0.8715  &  0.8599  &  0.8659  &  0.8605  &  0.8577 \\
5 &    0.5912 &   0.8183 &   0.8832 &   0.8664  &  0.8537  &  0.8610  &  0.8596  &  0.8538 \\
6 &    0.5724 &   0.8056 &   0.8742 &   0.8591  &  0.8497  &  0.8571  &  0.8566  &  0.8537 \\
7 &    0.5629 &   0.7983 &   0.8689 &   0.8580  &  0.8474  &  0.8551  &  0.8546  &  0.8508 \\
8 &    0.5477 &   0.7889 &   0.8623 &   0.8523  &  0.8433  &  0.8525  &  0.8518  &  0.8496 \\
\bottomrule
\end{tabular}
\end{center}
\end{table}

According to Table \ref{tab:mungA_res}, the calculated segregation degree is also influenced by the number of rows and columns the picture is splintered into. When number of parts is small, the value varies dramatically. However, the value goes stable when number of parts is big enough.

According to the table, the value of $Rows$ and $Cols$ also has considerable influence on the result. When the values of the rows or columns are too small, the calculated value varies significantly.  When the values of rows and columns are big enough, the calculated value becomes stable. 

When $Rows=3$ and $Cols=1$, the calculated segregation value is 0.6164. However, the calculated value is 1.0 when $Rows=1$ and $Cols=3$. In both cases, the picture is splintered into three parts. But the calculated segregation index varies significantly. This phenomenon indicates that segregation in horizontal direction differs from that in vertical direction. In order to get reliable segregation result, both $Rows$ and $Cols$ should be selected carefully. 

In Table \ref{tab:mungA_res}, when the value rows and columns is about $7$, the calculated result is stable and consistently.
Hence, the values of rows and column are set as $7$. Then, the calculated values are listed in Table \ref{tab:mung_res}.

\begin{table}
\scriptsize
\begin{center}
\caption{Segregation Result of Mungs}
\label{tab:mung_res}
\begin{tabular}{c|cccc}
\toprule
Sample Number & 1      & 2      & 3      & 4      \\
\hline
Segregation Index & 0.8546 & 0.8009 & 0.3030 & 0.2490 \\
\bottomrule
\end{tabular}
\end{center}
\end{table}

The segregation index values of mung samples are listed in Table \ref{tab:mung_res}. The segregation index is $0.8546$ when the mungs are squeezed together, as shown in Figure \ref{fig:mung1}. However, the segregation index decreased to $0.2490$ when the mungs scatter uniformly in the view port, the corresponding picture is shown in Figure \ref{fig:mung4}. When the mungs distribute more equally, the segregation index is lower. The results coincide with the distribution pattern of the mungs. 

\begin{figure} 
\centering 
\begin{subfigure}[b]{0.30\textwidth} \includegraphics[width=\textwidth]{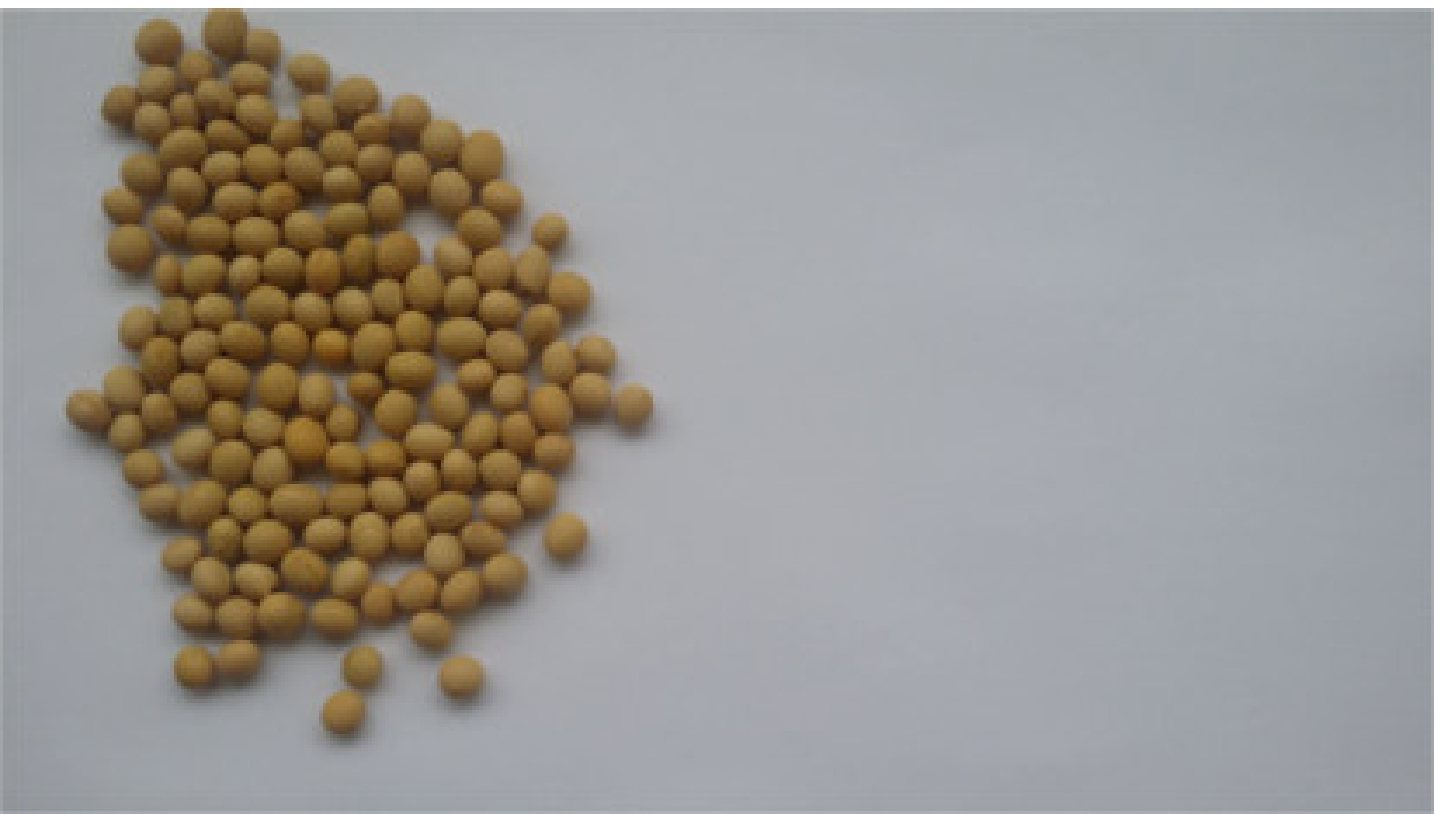} \caption{Sample $1$} \label{fig:soybean_A} \end{subfigure} ~ 
\begin{subfigure}[b]{0.30\textwidth} \includegraphics[width=\textwidth]{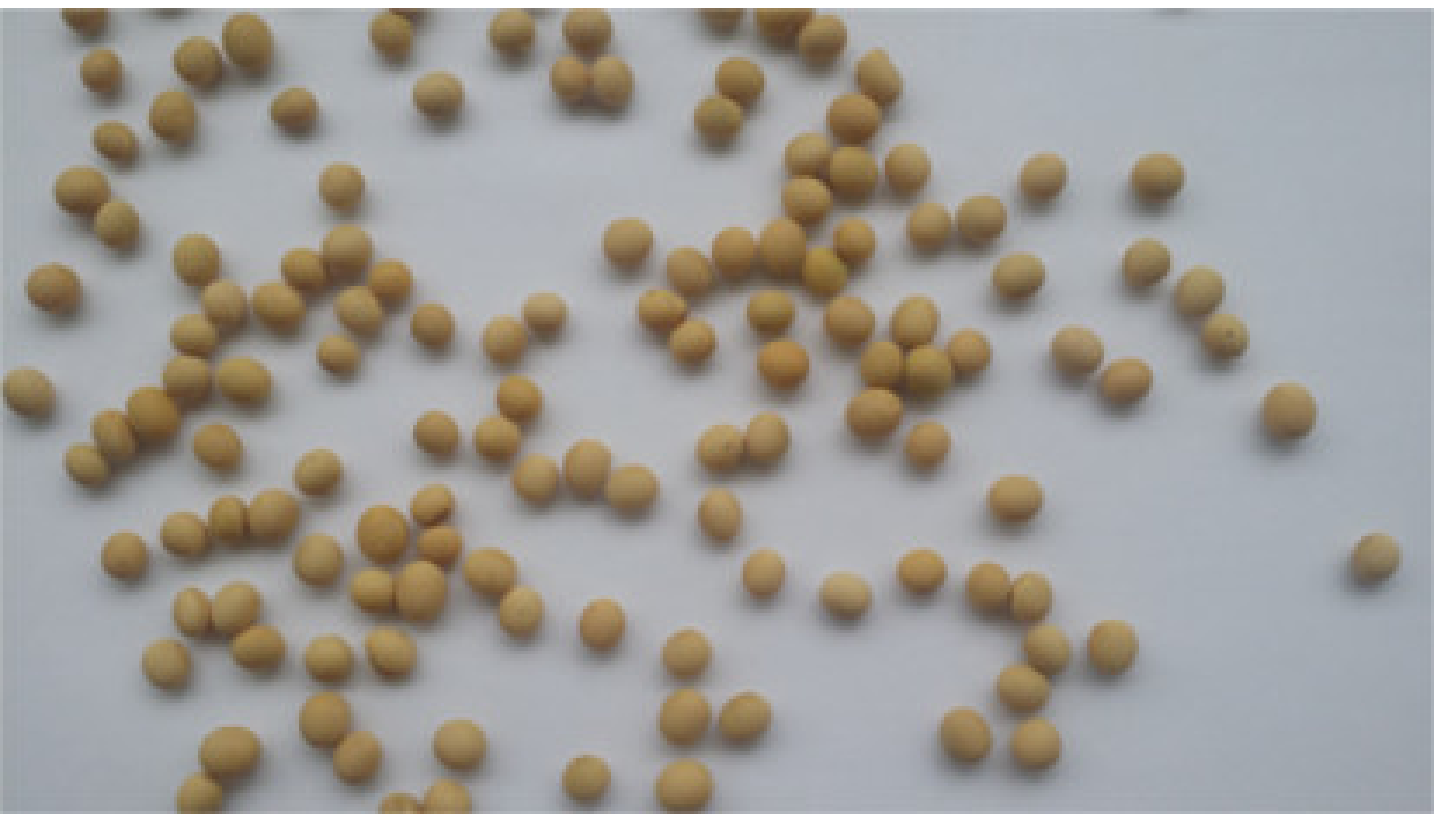} \caption{Sample $2$} \label{fig:soybean_B} \end{subfigure} ~ 
\begin{subfigure}[b]{0.30\textwidth} \includegraphics[width=\textwidth]{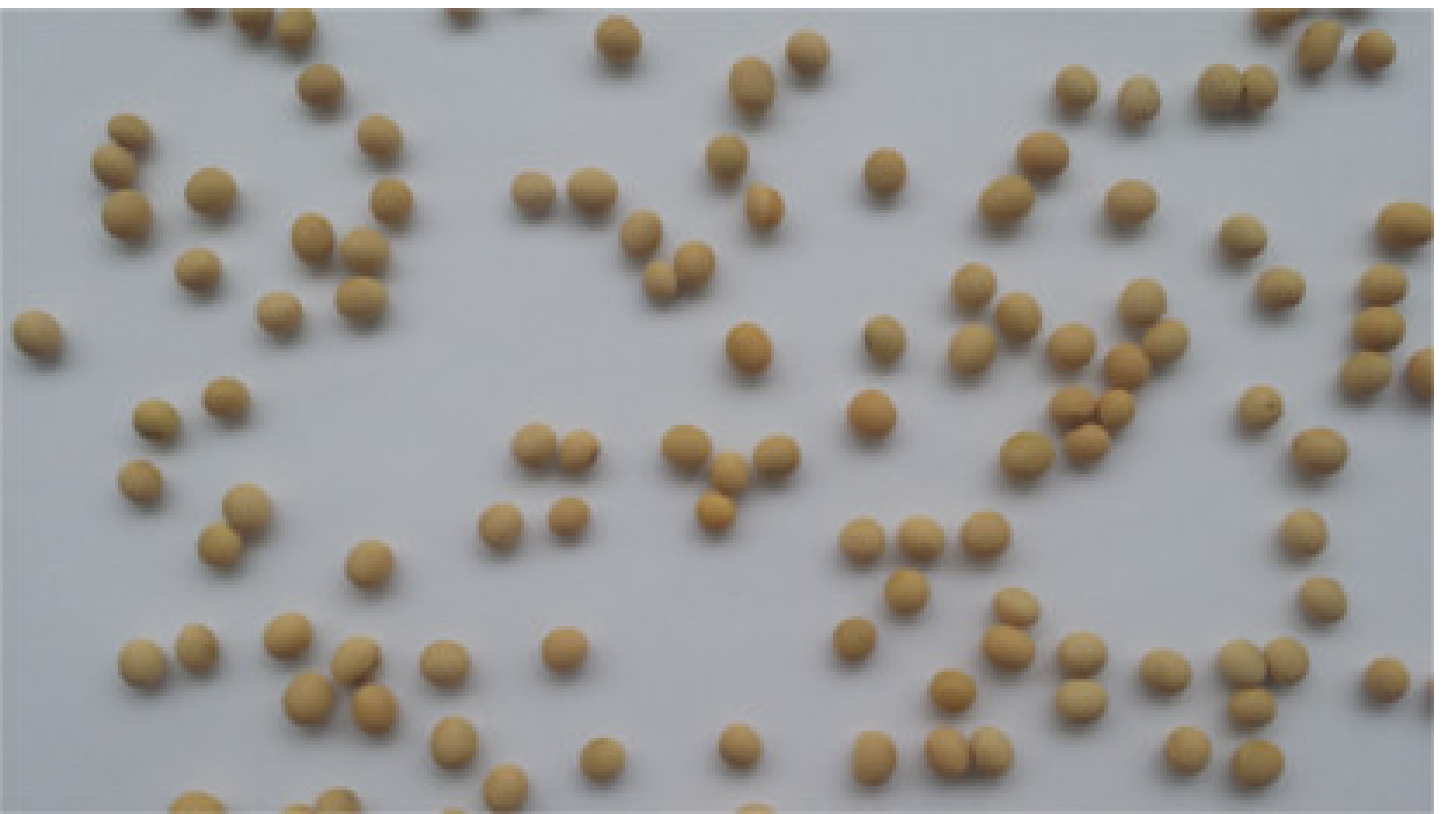} \caption{Sample $3$} \label{fig:soybean_C} \end{subfigure}
\caption{Distribution of Soybeans}\label{fig:soybean} 
\end{figure}

\begin{table}
\scriptsize
\begin{center}
\caption{Segregation Result of soybeans}
\label{tab:soybean_res}
\begin{tabular}{c|ccc}
\toprule
Sample Number     & 1      & 2      & 3       \\
\hline
Segregation Index & 0.7616 & 0.3716 & 0.3054  \\
\bottomrule
\end{tabular}
\end{center}
\end{table}

In the case of relative bigger soybeans, the distribution patterns of them are shown in Figure \ref{fig:soybean}. The calculate segregation index varies according to the same principle.
It decreases when soybeans distribute equally in the view port.  The values are shown in Table \ref{tab:soybean_res}.

\begin{figure} 
\centering 
\begin{subfigure}[b]{0.30\textwidth} \includegraphics[width=\textwidth]{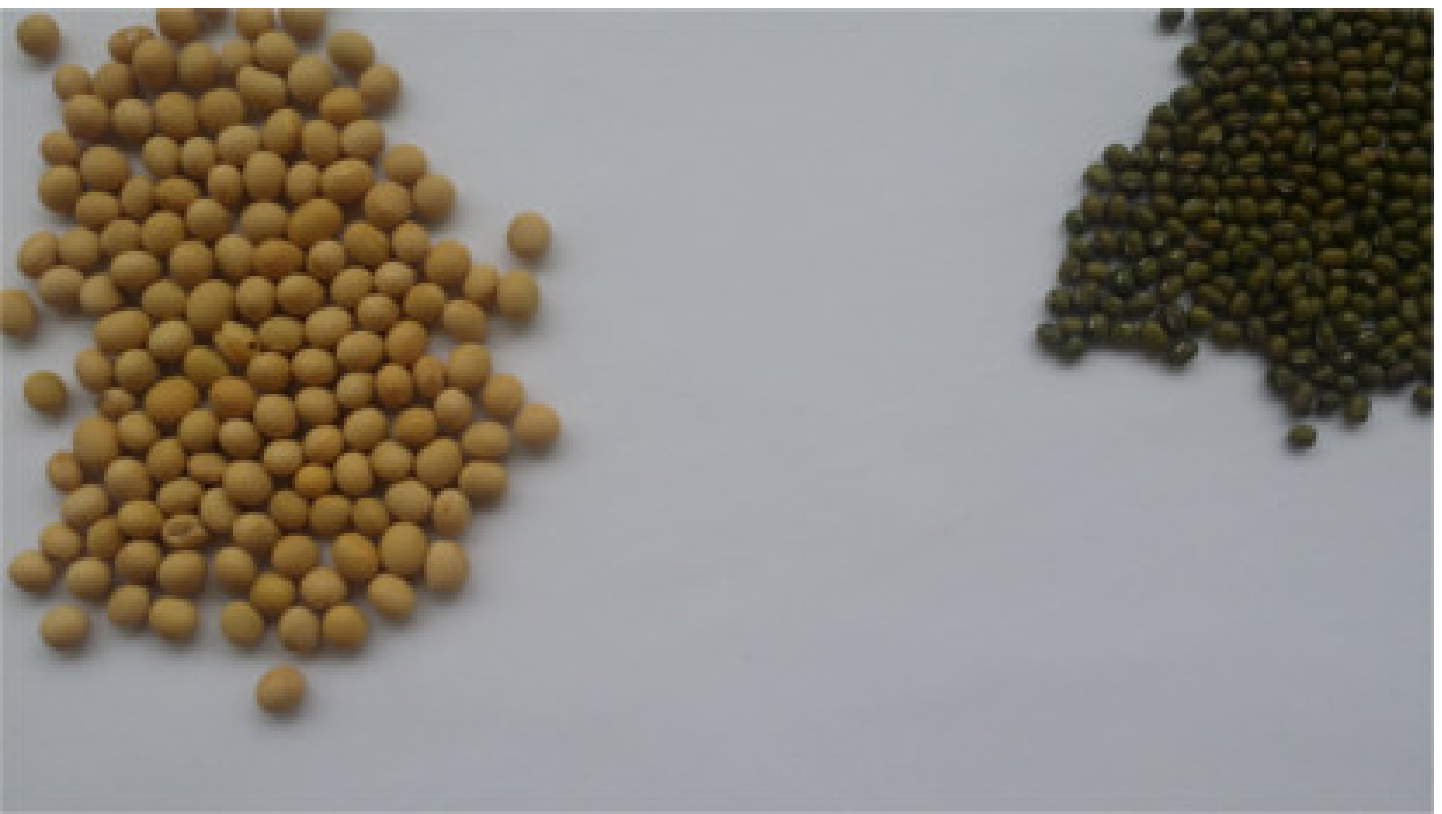} \caption{Sample $1$} \label{fig:sample_A} \end{subfigure} ~ 
\begin{subfigure}[b]{0.30\textwidth} \includegraphics[width=\textwidth]{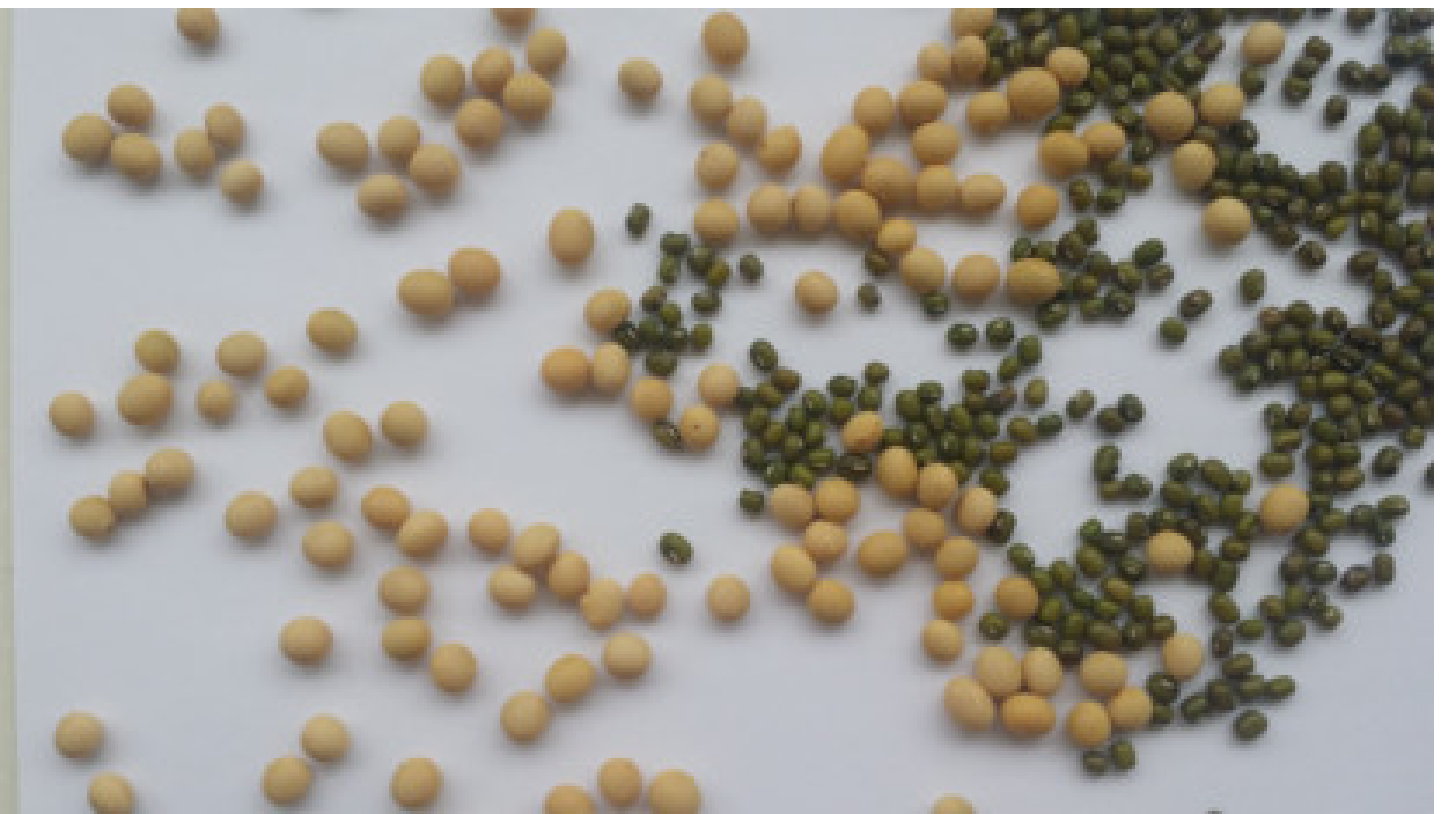} \caption{Sample $2$} \label{fig:sample_B} \end{subfigure} ~ 
\begin{subfigure}[b]{0.30\textwidth} \includegraphics[width=\textwidth]{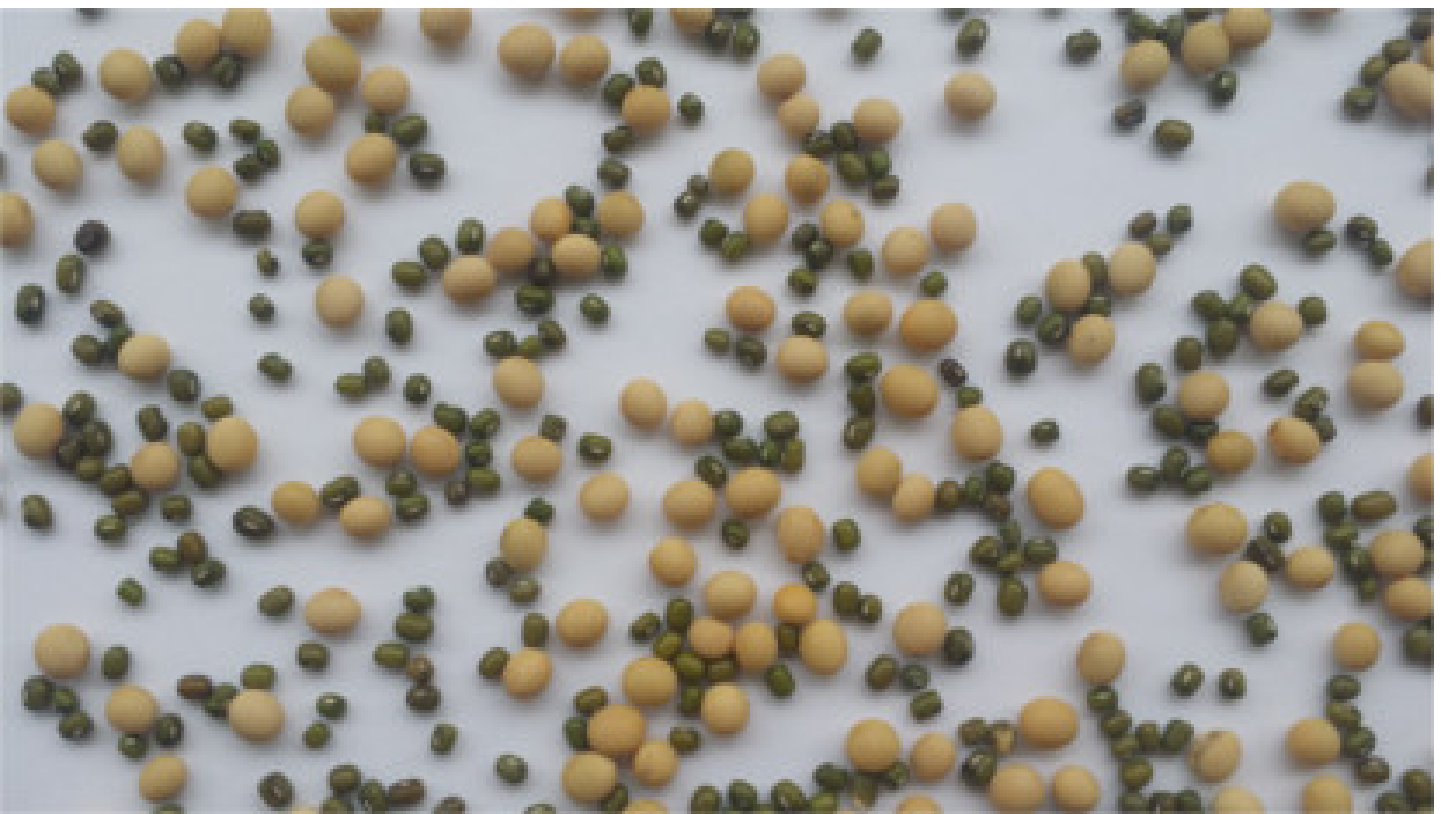} \caption{Sample $3$} \label{fig:sample_C} \end{subfigure}
\caption{Distribution of Samples}\label{fig:sample} 
\end{figure}

\begin{table}
\scriptsize
\begin{center}
\caption{Segregation Result of Samples}
\label{tab:sample_res}
\begin{tabular}{c|ccc}
\toprule
Sample Number     & 1      & 2      & 3       \\
\hline
Segregation Index & 0.6478 & 0.3933 & 0.1541  \\
\bottomrule
\end{tabular}
\end{center}
\end{table}

In Figure \ref{fig:sample}, mungs and soybeans distribute both in the pictures. When mungs and soybeans are separated clearly in Figure \ref{fig:sample_A}, the corresponding segregation index is $0.6478$. The segregation index is $0.1541$ when mungs and soybeans are mixed together. According to the results shown in Table \ref{tab:sample_res}, the proposed segregation index is suitable for segregation evaluation of different particles.

\begin{figure} 
\centering 
\begin{subfigure}[b]{0.30\textwidth} \includegraphics[width=\textwidth]{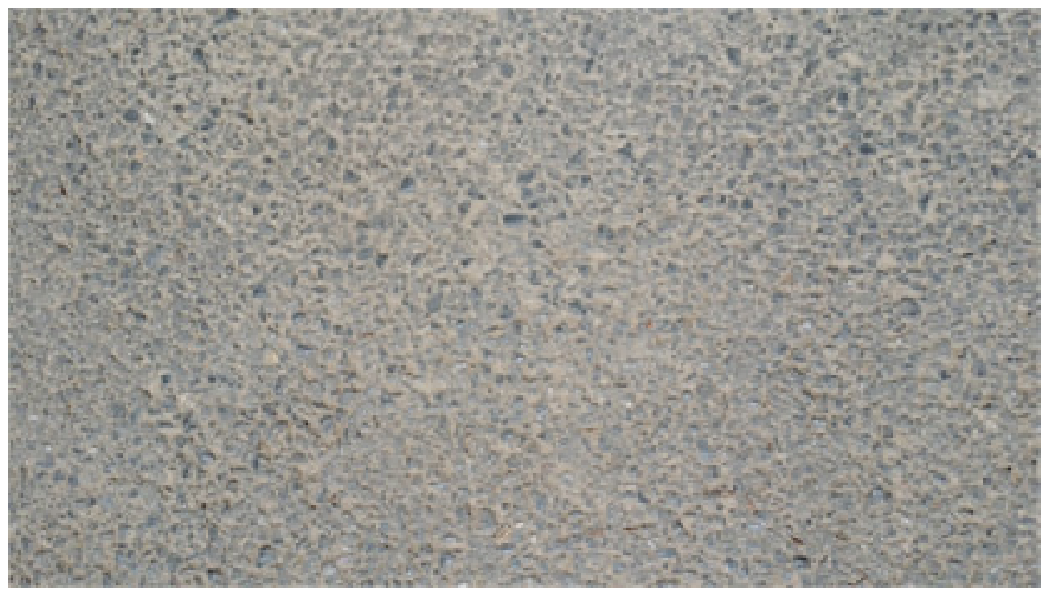} \caption{Origin Picture} \label{fig:pavement_origin} \end{subfigure} ~ 
\begin{subfigure}[b]{0.30\textwidth} \includegraphics[width=\textwidth]{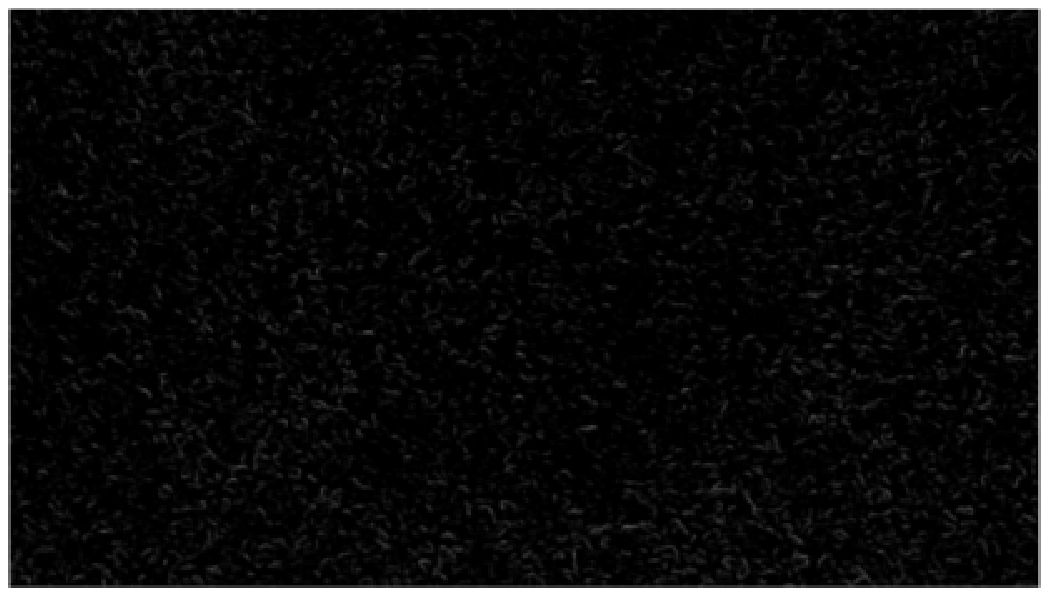} \caption{Edge Detect Result} \label{fig:pavement_edge} \end{subfigure} ~ 
\begin{subfigure}[b]{0.30\textwidth} \includegraphics[width=\textwidth]{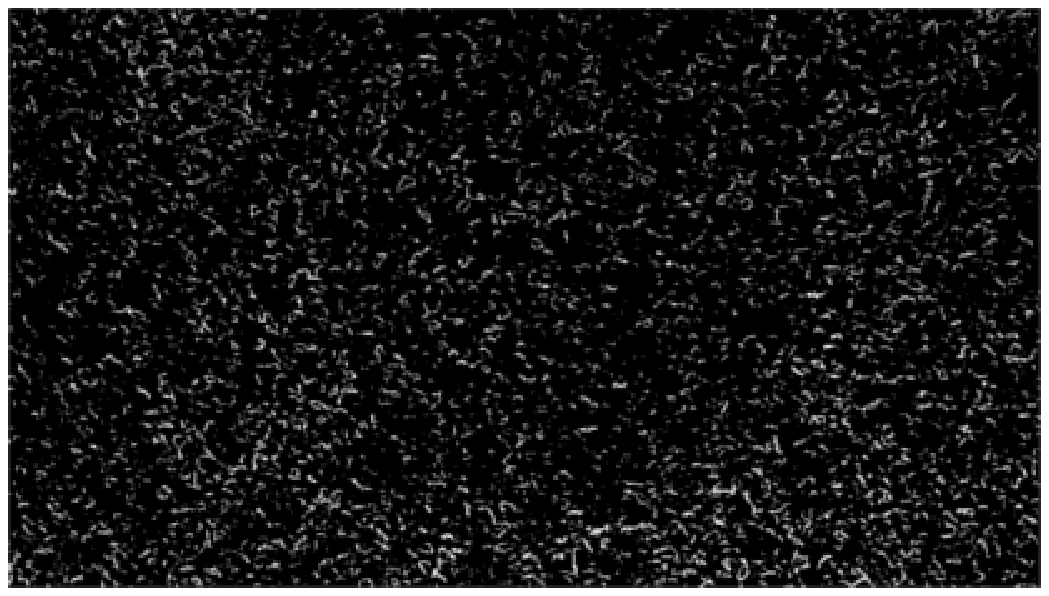} \caption{Binary Image} \label{fig:pavement_edge_bw} \end{subfigure}
\caption{Asphalt Pavement}\label{fig:pavement} 
\end{figure}

Segregation has considerable influence on asphalt pavement performance. Segregation evaluation of the particles in asphalt pavement is meaningful in practice.
The picture of constructed asphalt pavement is shown in Figure \ref{fig:pavement_origin}. The extracted edges of the particles are shown in Figure \ref{fig:pavment_edge}. 
The result picture is converted to binary image, which is shown in Figure \ref{fig:pavement_edge_bw}. Then, segregation index is computed according to Algorithm \ref{alg:segregation}. 
The results are tabulated in Table \ref{tab:pavement_res}. When $Rows=7$ and $Cols=7$, the segregation index is $0.1781$. It means segregation in the asphalt pavement sample is not significant.

\begin{table}
\scriptsize
\begin{center}
\caption{Segregation Result of Pavement}
\label{tab:pavement_res}
\begin{tabular}{c|cccccccc}
\toprule
\diagbox{Rows}{Cols} & 1 & 2 & 3 & 4 & 5 & 6 & 7 & 8  \\
\hline
   1 &   0.0000 &   0.0164 &   0.0622 &   0.0438 &   0.0487 &   0.0679 &   0.0723 &   0.0726 \\
   2 &   0.1908 &   0.1425 &   0.1425 &   0.1358 &   0.1339 &   0.1387 &   0.1407 &   0.1425 \\
   3 &   0.1682 &   0.1464 &   0.1477 &   0.1542 &   0.1498 &   0.1520 &   0.1548 &   0.1596 \\
   4 &   0.1665 &   0.1510 &   0.1563 &   0.1685 &   0.1572 &   0.1623 &   0.1682 &   0.1751 \\
   5 &   0.1630 &   0.1512 &   0.1630 &   0.1718 &   0.1656 &   0.1735 &   0.1751 &   0.1812 \\
   6 &   0.1595 &   0.1519 &   0.1605 &   0.1691 &   0.1645 &   0.1710 &   0.1753 &   0.1821 \\
   7 &   0.1538 &   0.1477 &   0.1569 &   0.1710 &   0.1667 &   0.1738 &   0.1781 &   0.1852 \\
   8 &   0.1519 &   0.1475 &   0.1586 &   0.1704 &   0.1653 &   0.1717 &   0.1783 &   0.1872 \\
\bottomrule
\end{tabular}
\end{center}
\end{table}

\subsection{Discussion}

According to the results, the proposed method is an effective approach to evaluate segregation between particles.

The number of parts of the picture splintered should be selected carefully in order to get reasonable results. Unduly high or low value of the parts number leads to unsharpness results. 
According to the experimental results, the picture is splintered into $7 \times 7$ parts is a good choice.

\section{Conclusion and Future Works}

Segregation has negative influence on material performance. In this work, an objective segregation measuring method is proposed, which can be implemented as automated system. In order to evaluate segregation degree qualitatively, digital picture of the particles are taken. Then, edges of the particles are extracted. The result picture is splintered to equal parts, average length of the edges in each parts is calculated. Segregation index is computed according to the edge length of the parts. The results show that the calculated segregation index coincides with intuition. 

In practice, the parameters such as relative size of the picture and particles should be adjusted in order to get meaningful and stable result.
Because the proposed segregation quantitative evaluation process is easy to implement as automated program, it would be a promising to ensure material producing process where segregation is important to material performance. This should be conducted in the future.

\section*{References}

\bibliography{AggregateSegregate}

\end{document}